\pdfoutput=1

\documentclass[twoside]{article}
\usepackage[accepted]{aistats2019}
\usepackage[round]{natbib}

\usepackage[utf8]{inputenc} 
\usepackage[T1]{fontenc}    
\usepackage{hyperref}       
\usepackage{url}            
\usepackage{booktabs}       
\usepackage{amsfonts}       
\usepackage{nicefrac}       
\usepackage{microtype}      

\usepackage[page]{appendix}

\RequirePackage{epsfig}
\RequirePackage{amssymb}
\RequirePackage{natbib}
\RequirePackage{graphicx}

\usepackage{amsthm}

\newtheorem{theorem}{Theorem}
 
\newtheorem{proposition}[theorem]{Proposition} 
\newtheorem{remark}[theorem]{Remark}
\newtheorem{corollary}[theorem]{Corollary}

\usepackage{epstopdf}
\usepackage{graphicx}
\usepackage{subcaption}

\usepackage{amsmath}

\usepackage{bbm}
\usepackage{bm}

\usepackage{algorithm}
\usepackage{algpseudocode}
\usepackage{mathtools}

\newcommand{\bfi}{\begin{fig}}
	\newcommand{\efi}{\end{fig}}
\newcommand{\btab}{\begin{tab}}
	\newcommand{\etab}{\end{tab}}
\newcommand{\barr}{\begin{array}}
	\newcommand{\earr}{\end{array}}
\newcommand{\beqq}{\begin{equation}}
\newcommand{\eeqq}{\end{equation}}
\newcommand{\beao}{\begin{eqnarray*}}
	\newcommand{\eeao}{\end{eqnarray*}\noindent}
\newcommand{\beam}{\begin{eqnarray}}
\newcommand{\eeam}{\end{eqnarray}\noindent}
\newcommand{\bdis}{\begin{displaymath}}
\newcommand{\edis}{\end{displaymath}\noindent}

\newcommand{\eps}{{\epsilon}}

\DeclareMathOperator{\e}{e} 
\newcommand{\E}{\mathop{\mathbb E}}

\begin{document}

%

%

\twocolumn[

\aistatstitle{Scalable Bayesian Learning for State Space Models using Variational Inference with SMC Samplers}

\aistatsauthor{ Marcel Hirt \And Petros Dellaportas  }
\aistatsaddress{University College of London, UK\\ \And  
	University College of London, UK,\\
	Athens University of Economics and Business, Greece	\\
	and The Alan Turing Institute, UK  } 

]


\begin{abstract}
	We present a scalable approach to performing approximate fully Bayesian inference in generic state space models. The proposed method is an alternative to particle MCMC that provides fully Bayesian inference of both the dynamic latent states and the static parameters of the model. We build up on recent advances in computational statistics that combine variational methods with sequential Monte Carlo sampling and we demonstrate the advantages of performing  full Bayesian inference over the static parameters rather than just performing variational EM approximations. We illustrate how our approach enables scalable inference in multivariate stochastic volatility models and self-exciting point process models that allow for flexible dynamics in the latent intensity function. 
\end{abstract}

\section{Introduction}

We deal with generic state-space models (SSM) which may be nonlinear and non-Gaussian.  Inference for this important and popular family of statistical models presents tremendous challenges that has prohibited their widespread applicability.  The key difficulty is that inference on the latent process of the model depends crucially on unknown static parameters that need to be also estimated.  While MCMC samplers are unsatisfactory because they both  fail to produce high dimensional, efficiently mixing Markov chains and because they are inappropriate for on-line inference, sequential Monte Carlo (SMC) methods \citep{kantas2015particle} provide the tools to construct successful viable implementation strategies.  In particular, particle MCMC \citep{andrieu2010particle} utilises SMC to build generic efficient MCMC algorithms that provide inferences for both static parameters and latent paths.  We provide a scalable alternative to these methods via an approximation that combines SMC  and variational inference.

We introduce a new variational distribution that unlike recent strand of literature \citep{maddison2017filtering,naesseth2017variational,le2017auto} performs variational inference also on the static parameters of the SSM. This is essential for various reasons. First, when there is dependency between static and dynamic parameters posterior inference may be inaccurate if the joint posterior density is approximated by conditioning on fixed values of static parameters. Second, inferring the static parameter is often the primary problem of interest: for example, for biochemical networks and models involving Lotka Voltera equations, we are not interested in the population of the species per se, but we want to infer some chemical rate constants (such as reaction rates or predation/growth rates), which are parameters of the transition density;  in neuroscience, Bayesian decoding of neural spike trains is often made via a state-space representation of point processes in which inference for static parameters is of great importance.
Finally, for complex dynamic systems it is often advisable to improve model compression or interpretability by encouraging sparsity and such operations may require inference for the posterior densities of the static parameters.

Sampling from the new variational distribution involves running a SMC algorithm which yields an unbiased estimate of the likelihood for a fixed static parameter value. Importantly, we show that the SMC algorithm constructs a computational graph that allows for optimisation of the variational bound using stochastic gradient descent. We provide some empirical evidence that variational inference on static parameters can give better predictive performance, either out-of sample in the linear Gaussian state space  model or in-sample for predictive distributions in a multivariate stochastic volatility model.  We also illustrate our method by modelling fairly general intensity functions in a multivariate Hawkes process model.

\section{Background}	

Let us begin by introducing the standard inference problem in a generic SSM, followed by a review of the SMC approach to sample from a sequence of distributions arising in such probabilistic structures. SSMs are characterized by a latent Markov state process $\{X_n\}_{n\geq 0}$ on $\mathbb{R}^{d_x}$ and an observable process $\{Y_n\}_{n\geq 0}$ on $\mathbb{R}^{d_y}$. We follow the standard convention of using capital letters for random variables and the corresponding lower case letter to denote their values. The dynamics of the latent states is determined, conditional on a static parameter vector $\theta \in \Theta$, by a transition probability density 
\[X_{n}| (\theta, X_{n-1}=x_{n-1},Y_{n-1}=y_{n-1}) \sim f_{\theta}(\cdot | x_{n-1},y_{n-1}), \]	
along with an initial density $X_0  \sim f_{\theta}(\cdot)$. The observations are assumed to be conditionally iid given the states with density given by
\[ Y_n | (\theta, X_{0:n}=x_{0:n},Y_{0:n-1}=y_{0:n-1})\sim g_{\theta}(\cdot | x_n), \]
for any $n\geq 0$ with the generic notation $x_{0:n}=(x_0,...,x_n)$. 

We consider a Bayesian framework and assume $\theta$ has a prior density $p(\theta)$. Consequently, for observed data $y_{0:M}$, we perform inference using the posterior density 
\beqq \pi(\theta,x_{0:M})\coloneqq p(\theta,x_{0:M}| y_{0:M})\propto p(\theta) p_{\theta}(x_{0:M},y_{0:M}),\label{posterior}\eeqq
where the joint density of the latent states
and observations given a fixed static parameter value $\theta$ writes as
\begin{align} &p_{\theta}(x_{0:M},y_{0:M})=\gamma_{\theta}(x_{0:M}) \nonumber\\
\coloneqq &f_{\theta}(x_0) \prod_{n=1}^M f_{\theta}(x_n| x_{n-1},y_{n-1})\prod_{n=0}^Mg_{\theta}(y_n| x_n).\label{joint_density}
\end{align}
The posterior density $p(\theta,x_{0:M}| y_{0:M})$ is in general intractable, as is
\beqq p_{\theta}(x_{0:M}| y_{0:M})=\frac{\gamma_{\theta}(x_{0:M})}{p_{\theta}(y_{0:M})},\label{filtering_density_pf} \eeqq
where $p_{\theta}(y_{0:M})=\int p_{\theta}(x_{0:M},y_{0:M})dx_{0:M}$.
However, an SMC algorithm can be used to approximate $(\ref{filtering_density_pf})$. A brief review of how this sampling algorithm proceeds is as follows and further details can be found in \citet{doucet2000sequential,doucet2009tutorial}.\\ 
SMC methods approximate $p_{\theta}(x_{0:n}| y_{0:n})$ using a set of $K$ weighted random samples $X_{0:n}^{1:K}=(X_{0:n}^1,...,X_{0:n}^K)$, also called particles, having positive weights $W_n=W_n^{1:K}$, so that $p_{\theta}(x_{0:n}| y_{0:n})\approx \hat{p}_{\theta}(x_{0:n}| y_{0:n})=\sum_{k=1}^K W_{n}^k \delta_{X_{0:n}^k}(x_{0:{n}})$. Here, $\delta$ denotes the Dirac delta function. To do so, one starts at $n=0$ by sampling $X_0^k$ from an importance density $M_0^{\phi}(\cdot | y_0)$, parametrized with $\phi$, where $\phi$ can depend on the static parameters $\theta$. For any $n\geq 1$, we first resample an ancestor variable $A_{n-1}^k$ that represents the 'parent' of particle $X_{0:n}^k$ according to $A_{n-1}^k\sim r(\cdot | W_{n-1})$, where $r$ is a categorical distribution on $\{1,...,K\}$ with probabilities $W_{n-1}$. We then set $W_{n-1}=\frac{1}{K}$ and proceed by extending the path of each particle by sampling from a transition kernel $X_n^k\sim M_n^{\phi}(\cdot | y_n,X_{0:n-1}^{A_{n-1}^k})$. This yields an updated latent path $X_{0:n}^k=(X_{0:n-1}^{A_{n-1}^k},X_n^k)$ for which we compute the incremental importance weight

\[\alpha_n(X_{0:n}^k)=\frac{\gamma_{\theta}(X_{0:n}^k)}{\gamma_{\theta}(X_{0:n-1}^k)M_n^{\phi}(X_n^k | y_n, X_{0:n-1}^{A_{n-1}^k})}.\]
We set $w_n(X_{0:n}^k)=W_{n-1}^k \alpha_n(X_{0:n}^k)$ as well as
$W_n^k=\frac{w_n(X_{0:n}^k)}{\sum_l w_n(X_{0:n}^l)}$ and define 
\[\hat{Z}_n^{\theta,\phi}\coloneqq \prod_{m=0}^n\sum_{k=1}^Kw_m(X_{0:m}^k),\]
which is an unbiased and strongly consistent estimator of $p_{\theta}(y_{0:n})$, see \citet{del1996non}. A pseudo-code (Algorithm \ref{SMCAlgo}) for this standard SMC sampler can be found in Appendix \ref{AppendixSmcSampler}. It is possible to perform the resampling step only if some condition on $W_{n-1}$ is satisfied, see Algorithm \ref{SMCAlgo}. For simplicity, we assume that the particles are resampled at every step. The density of all variables generated by this SMC sampler for a fixed static parameter value $\theta$ is given by
\begin{align*}&q_{\phi}(x_{0:M}^{1:K},a_{0:M-1}^{1:K},l| \theta)
=w_M^l\prod_{k=1}^K M_0^{\phi}(x_0^k| y_0) \\& \cdot \prod_{n=1}^M \prod_{k=1}^K r(a_{n-1}^k | w_{n-1})M_n^{\phi}({x_n^k} | y_n, x_{0:n-1}^{a_{n-1}^k}), 
\end{align*}
where $l$ is a final particle index drawn from a categorical distribution with weights $W_M$.
Since $\hat{Z}_n^{\theta, \phi}$ is unbiased, we have 
\beqq {\E}_{q_{\phi}(x_{0:M}^{1:K},a_{0:M-1}^{1:K},l| \theta)}\left[ \hat{Z}_M^{\theta,\phi}\right]=p_{\theta}(y_{0:M}).\label{SMC_unbiased}\eeqq

\section{Variational bounds for state space models using SMC samplers}

Variational inference \citep{jordan1999introduction,wainwright2008graphical,blei2017variational} allows Bayesian inference to scale to large data sets \citep{hoffman2013stochastic} and is applicable to a wide range of models \citep{ranganath2014black,kucukelbir2017automatic}. It generally postulates a family of approximating distributions with variational parameters that minimize some divergence, most commonly the KL divergence, between the approximating distribution and the posterior. The quality of the approximation hinges on the expressiveness of the variational family.

Let $q_{\psi}(\theta)$ be a distribution on $\Theta$ with variational parameters $\psi$. We aim to approximate the posterior density $p(\theta,x_{0:M}| y_{0:M})$ in $(\ref{posterior})$  with a variational distribution that results as an appropriate marginal of auxiliary variables arising from an SMC sampler of the form
\beqq q_{\psi,\phi}(\theta,x_{0:M}^{1:K},a_{0:M-1}^{1:K},l)\coloneqq q_{\psi}(\theta)q_{\phi}(x_{0:M}^{1:K},a_{0:M-1}^{1:K},l| \theta),\label{variational_distribution}\eeqq
defined precisely below. Note that sampling from the extended variational distribution (\ref{variational_distribution}) just means sampling $\theta\sim q_{\psi}(\theta)$ and then running a particle filter using the sampled value $\theta$ as the static parameter. 

We introduce the proposed variational bound first as a lower bound on $\log p(y_{0:M})-\text{KL}(q_{\psi}(\theta)||p(\theta | y_{0:M}))$. We then show that optimizing the proposed bound means minimizing the KL-divergence between the extended variational distribution $(\ref{variational_distribution})$ and an extended target density that resembles closely the density targeted in particle MCMC methods.

We can write $p(\theta | y_{0:M})=p(\theta)p_{\theta}(y_{0:M})/p(y_{0:M})$. Hence, using the fact that the likelihood estimator is unbiased $(\ref{SMC_unbiased})$ and due to Jensen's inequality,
\begin{align*}
	&-\text{KL}(q_{\psi}(\theta)||p(\theta | y_{0:M})) +\log p(y_{0:M})\\
	=&{\E}_{q_{\psi}(\theta)}\left[\log p_{\theta}(y_{0:M})+\log p(\theta) - \log q_{\psi}(\theta)\right] \\
	=&{\E}_{q_{\psi}(\theta)}\left[\log {\E}_{q_{\phi}(x_{0:M}^{1:K},a_{0:M-1}^{1:K},l| \theta)}\left[\hat{Z}_M^{\theta,\phi}\right] +\log \frac{p(\theta)}{q_{\psi}(\theta)}\right]\\
	\geq & {\E}_{q_{\psi}(\theta)}\left[ {\E}_{q_{\phi}(x_{0:M}^{1:K},a_{0:M-1}^{1:K},l| \theta)}\left[\log \hat{Z}_M^{\theta,\phi}\right] +\log \frac{p(\theta)}{ q_{\psi}(\theta)}\right]\\
	=:& \mathcal{L}(\psi,\phi).
\end{align*}
In particular, $\mathcal{L}(\psi,\phi)$ is a lower bound on $p(y_{0:M})- \text{KL}(q_{\psi}(\theta)||p(\theta | y_{0:M}))$.

\begin{remark}[\bfseries Inference for multiple independent time series]
	Instead of considering one latent process $\{X\}$ and observable process $\{Y\}$, we can also consider $S$ independent latent processes $\{X^s\}_{s=1,...,S}$ with corresponding observable processes $\{Y^s\}_{s=1,...,S}$ described by the same static parameter $\theta$. We obtain a lower bound on $p(y_{0:M})-\text{KL}(q_{\psi}(\theta)||p(\theta | y^1_{0:M},..., y^S_{0:M}))$ given by
	\begin{align*} 
	&{\E}_{q_{\psi}(\theta)}\Bigg[ {\E}_{\prod_sq_{\phi}(x_{0:M}^{s,1:K},a_{0:M-1}^{s,1:K},l^s| \theta)}\left[\sum_{s=1}^S\log \hat{Z}_{M,s}^{\theta,\phi}\right] \\&+\log p(\theta) - \log q_{\psi}(\theta)\Bigg],
	\end{align*}
	where $\hat{Z}^{\theta,\phi}_{M,s}$ is the estimator of $p_{\theta}(y_{0:M}^s)$. Note that we can obtain an unbiased estimate of this bound by sampling an element $s\in\{1,...,S\}$ and using $S\cdot \log \hat{Z}_{M,s}^{\theta,\phi}$ as an estimate of $\sum_{s'=1}^S\log \hat{Z}_{M,s'}^{\theta,\phi}$, thereby allowing our method to scale to a large number of independent time series. For ease of exposition, we formulate our results for a single time series only.
\end{remark}
Next, we show that the variational bound can be represented as the difference between the log-evidence and the KL divergence between the variational distribution and an extended target density. More concretely, following \citet{andrieu2010particle}, we consider a target density on the extended space $ \Theta \times \mathcal{X}$, $\mathcal{X} \coloneqq (\mathbb{R}^{d_x})^{(M+1)K}\times \{1,...,K\}^{MK+1}$,
\begin{align*}
&\tilde{\pi}(\theta,x_{0:M}^{1:K},a_{0:M-1}^{1:K},l)  \coloneqq \frac{\pi(\theta,x^l_{0:M})}{K^{M+1}} \\ &\cdot  \frac{q_{\phi}(x_{0:M}^{1:K},a_{0:M-1}^{1:K},l| \theta)}{M_0^{\phi}(x_0^{b_0^l}| y_0) \prod_{n=1}^M r(b_{n-1}^l| w_{n-1}) M_n^{\phi}(x_n^{b_n^l}| y_n, x_{0:n-1}^{b_{n-1}^l})}.
\end{align*}
Here, we have defined $b_M^l=l$ and $b_n^l=a_n^{b_{n+1}^l}$ for $n=M-1,...,1$, i.e. $b_n^l$ is the index that the ancestor of particle $X_{0:M}^l$ at generation $n$ had.
It follows, using $r(b_n^l| w_{n-1})=w_{n-1}^{b_{n-1}^l}$, that the ratio between the extended target density and the variational distribution is given by
\begin{align}	
&\frac{\tilde{\pi}(\theta,x_{0:M}^{1:K},a_{0:M-1}^{1:K},l)}{ q_{\phi,\psi}(\theta,x_{0:M}^{1:K},a_{0:M-1}^{1:K},l)}\nonumber \\
=&\frac{ 	
	K^{-(M+1)}p(\theta) p_{\theta}(x_{0:M}^l,y_{0:M})/p(y_{0:M})}
{q_{\psi}(\theta)W_M^lM_0^{\phi}(x_0^{b_0^l}| y_0)\prod_{n=1}^M W_{n-1}^{b_{n-1}^l}M_{n}^{\phi}(x_n^{b_n^l} | y_n,x_{0:n-1}^{b_{n-1}^l})}.\label{ratio_extended_target}
\end{align}

\begin{proposition}[\bfseries KL divergence in extended space]\label{proposition_extended_space}
	It holds that
	\[ \mathcal{L}(\psi,\phi)=-\text{KL}(q_{\psi,\phi}||\tilde{\pi})+\log p(y_{0:M}).\]	
\end{proposition}
The proof can be found in Appendix \ref{AppendixExtendedSpaceRepresentation}. Recall that we have introduced $\mathcal{L}(\psi,\phi)$ so that its maximisation pushes the variational approximation of the static parameter $\theta$ closer to its true posterior as measured by the KL divergence. The above proposition shows that this objective also minimizes the KL divergence between densities on an extended space that includes multiple latent paths. To elucidate further the relation between the variational distribution of a single latent path and its posterior, we need to introduce a further distribution. Consider the density under $\tilde{\pi}$ of the variables generated by a SMC algorithm conditional on a fixed latent path $(x_{0:M}^{l},b_{0:M-1}^{l})$. This is known as a conditional SMC algorithm \citep{andrieu2010particle}, with distribution given by

\begin{align*}
&\tilde{\pi}_{\text{CSMC}}(x_{0:M}^{\neg b_{0:M}^l},a_{0:M-1}^{\neg b_{0:M-1}^l} | \theta,x_{0:M}^l,b_{0:M}^l)\\
=&\frac{ q_{\phi}(x_{0:M}^{1:K},a_{0:M-1}^{1:K},l| \theta )}{W_M^lM_0^{\phi}(X_0^{b_0^l}|y_0)\prod\limits_{n=1}^M r(b_{n-1}^l |W_{n-1})M_{n}^{\phi}(x_n^{b_n^l} | y_n, x_{0:n-1}^{b_{n-1}^l})},
\end{align*}

where $\neg b_{0:M}^l$ are the indices of all particles that are not equal to $b_{0:M}^l$. We obtain the following corollary proved in Appendix \ref{AppendixMarginalKL}. 
\begin{corollary}[\bfseries Marginal KL divergence and marginal ELBO]\label{corollary_marginal_bound}
	The KL divergence in the extended space is an upper bound on the KL divergence between the marginal variational approximation and the posterior, with the gap between bounds being
	\begin{align*}&\text{KL}\left(q_{\psi,\phi}(\theta,x_{0:M}^{1:K},a_{0:M-1}^{1:K},l)||\tilde{\pi}(\theta,x_{0:M}^{1:K},a_{0:M-1}^{1:K},l)\right)\\
	&-\text{KL}\left(q_{\psi,\phi}(\theta,x_{0:M})||\pi(\theta,x_{0:M}^l)\right)\\
		=&{\E}_{q_{\psi,\phi}(\theta,x_{0:M}^l,b_{0:M}^l)}\bigg[ \\
		& \quad \text{KL}(q_{\phi}(x_{0:M}^{\neg b_{0:M}^l},a_{0:M-1}^{\neg b_{0:M-1}^l})| \theta,x_{0:M}^l,b_{0:M}^l) || \\
		& \quad \quad \tilde{\pi}_{\text{CSMC}}(x_{0:M}^{\neg b_{0:M}^l},a_{0:M-1}^{\neg b_{0:M-1}^l} | \theta,x_{0:M}^l,b_{0:M}^l))\bigg].
	\end{align*}
	Particularly, $\mathcal{L}$ is a lower bound compared to the standard ELBO using the marginal $q_{\psi,\phi}(\theta,x_{0:M})$ with $x^l_{0:M}=x_{0:M}$ as the variational distribution:
	\[\mathcal{L}(\psi,\phi)\leq -\text{KL}\left(q_{\psi,\phi}(\theta,x_{0:M})||\pi(\theta,x_{0:M})\right)+\log p(y_{0:M}).\]
	
\end{corollary}

The proposed surrogate objective resembles variational bounds with auxiliary variables \citep{salimans2015markov,maaloe2016auxiliary,ranganath2016hierarchical} where the gap between the two bounds is expressed by the KL-divergence between the variational approximation of the auxiliary variable given the latent variable of interest and a so-called reverse model. Here, this reverse model is specified by the conditional SMC algorithm. 
The above corollary implies that the variational bound is looser than the standard ELBO with the auxiliary variables integrated out. This marginal variational distribution cannot in general be evaluated analytically. However, we can obtain unbiased estimates of it by computing the log-likelihood estimate under a conditional SMC algorithm, resembling a particle Gibbs update. This constitutes an extension of Proposition 1 in \citet{naesseth2017variational}. We present a proof in Appendix \ref{AppendixMarginalDist}. 

\begin{proposition}[\bfseries Marginal variational distribution]\label{proposition_marginal_distribution}
	We have
	\begin{align*}&q_{\psi,\phi}(\theta, x^l_{0:M},b_{0:M}^l)=q_{\psi}(\theta)\gamma_{\theta}(x^l_{0:M}) \\
	& \cdot {\E}_{\tilde{\pi}_{\text{CSMC}}(x_{0:M}^{\neg b_{0:M}^l},a_{0:M-1}^{\neg b_{0:M-1}^l} | \theta, x_{0:M}^l)}\left[\left({\hat{Z}^{\theta,\phi}_M}\right)^{-1}\right]
	\end{align*}
	and there exists $c(\theta,\phi)<\infty$ so that	
	\begin{align*}&\text{KL}(q_{\psi,\phi}(\theta,x_{0:M})||p(\theta,x_{0:M}|y_{0:M})\\\leq &{\E}_{q_{\psi}(\theta)}\left[\frac{c(\theta,\phi)}{K}\right]+\text{KL}(q_{\psi}(\theta)||p(\theta|y_{0:M})).
	\end{align*}
\end{proposition}
The last inequality in Proposition \ref{proposition_marginal_distribution} is a straightforward extension of an analogous result in the EM setting \citep{naesseth2017variational}. It implies that, for fixed variational parameters $\psi$ and $\phi$, the approximation becomes more accurate for increasing $K$. Sampling from this distribution can be seen as an extension of visualizing the expected importance weighted approximation in Importance Weighted Auto-Encoders \citep{cremer2017reinterpreting}. Since this distribution can be high-dimensional, the preceding proposition gives an alternative to kernel-density estimation.


Lastly, from a different angle, the variational objective can be seen as a sequential variational-autoencoding (VAE) bound. Indeed, as a consequence of Proposition \ref{proposition_extended_space} and equation (\ref{ratio_extended_target}), we obtain immediately the following result. We elaborate on it further in the next section.

\begin{corollary}[\bfseries Sequential VAE representation]\label{proposition_VAE_representation} The variational bound can be written as
	\begin{align*}\mathcal{L}(\psi,\phi)= {\E}_{q_{\psi}(\theta)}\Bigg[ &
		{\E}_{q_{\phi}(x_{0:M}^{1:K},a_{0:M-1}^{1:K},l| \theta)}\bigg[ \\
		 & \quad \sum_{n=0}^M \log g_{\theta}(y_n| x_n^{b_n^l})  -\log  W_n^{b_n^l} \\ 
		& \quad +\log \frac{ f_{\theta}(x_n^{b_n^l} | x_{n-1}^{b_{n-1}^l},y_{n-1})}{M^{\phi}(x_n^{b_n^l}| y_n, x_{0:n}^{b_{n-1}^l})}\bigg] \Bigg]\\
		& 
		-(M+1)\log K-\text{KL}(q_{\psi}(\theta)||p(\theta)).
	\end{align*}
	
\end{corollary}

\section{Related Work}
The representation in Corollary \ref{proposition_VAE_representation} allows us to contrast the variational bound to previously considered sequential VAE frameworks  \citep{chung2015recurrent,archer2015black, fraccaro2016sequential,krishnan2017structured,goyal2017z}. The introduced bound contains the cross-entropy between the proposal distribution and the likelihood common to sequential VAE bounds. However, this reconstruction error is only evaluated for surviving particles. Similarly, while a sequential VAE framework includes a KL-divergence between the proposal distribution and the prior transition probability, the log-ratio of these two densities is only evaluated for a surviving path. Most work using sequential VAEs have considered observation and state transition models parametrised by neural networks, and given the high-dimensionality of the static parameters, have confined their analysis to variational EM inferences. This is also the case for the approaches in \citet{maddison2017filtering,naesseth2017variational,le2017auto}, to which this work is most closely related. They have demonstrated that resampling increases the variational bound compared to a sequential IWAE \citep{burda2015importance} approach. \citet{rainforth2018tighter} demonstrated that increasing the number of particles leads to a worse signal to noise ratio of the gradient estimate of the proposal parameters in an IWAE setting. \citet{le2017auto} suggested to use fewer particles without resampling for calculating the proposal gradient. A possible approach left for future work would be to consider a different resampling threshold for the proposal gradients. Finally, the objective in this work differs from adaptive SMC approaches optimizing the reverse KL-divergence (or $\chi^2$-divergence) between the posterior and the proposal, cf. \citet{cornebise2008adaptive,gu2015neural}.

\section{Optimization of the variational bound}

The gradient of the variational bound is given by 
\begin{align} &\nabla_{\psi,\phi} \mathcal{L}(\psi,\phi) \label{grad_bound} \\
=&\nabla_{\psi,\phi} \left({\E}_{q_{\psi}(\theta)}\left[ {\E}_{q_{\phi}(x_{0:M}^{1:K},a_{0:M-1}^{1:K},l| \theta)}\left[\log \hat{Z}_M^{\theta,\phi}\right] \right] \right) \nonumber \\
& \phantom{=}+ \nabla_{\psi} \left( {\E}_{q_{\psi}(\theta)}\left[\log \frac{p(\theta)}{q_{\psi}(\theta)}\right] \right).\nonumber
\end{align}

We focus on the gradient of the first expectation and note that the gradient of the second expectation can be estimated by standard (black-box) approaches in variational inference, depending of course on the chosen variational approximation. If for instance the variational distribution over the static parameters is continuously reparametrisable, one can use standard low-variance reparametrised gradients \citep{kingma2014auto,rezende2014stochastic,titsias2014doubly}. This is the gradient estimator that we use in our experiments in combination with mean-field variational families. We assume that the proposals $X_n^k\sim M_n^{\phi}(\cdot | y_n, x_{0:n-1}^{a_{n-1}^k})$ are reparametrisable, i.e. there exists a differentiable deterministic function $h_{\phi}$ such that $X_n^k=h_{\phi}(X_{0:n-1}^{A_{n-1}^k},\eps_n^k)$, with $\eps_n^k\sim p(\cdot)$ continuous and independent of $\phi$. Similarly, we assume that the variational distribution of the static parameters is reparametrisable, i.e. there exists a differentiable deterministic function $h_{\psi}$ such that $\theta=h_{\psi}(\eta)$, with $\eta \sim p(\cdot)$ continuous and independent of $\psi$.
We abbreviate $\bm{\eps}=\eps_{0:M}^{1:K}$, $\bm{x}=x_{0:M}^{1:K}$ and $\bm{a}=a_{0:M-1}^{1:K}$. Using the product rule, observe that the first gradient in $(\ref{grad_bound})$ is

\begin{align*}
	& \nabla_{\psi,\phi} \int p(\eta) p(\bm{\eps})q_{\phi}( \bm{a} | \theta,\bm{x})\\
	 &\phantom{=}\cdot \log \hat{Z}_M^{\theta,\phi}  d(\eta, \bm{a},\bm{\eps}) \bigg|_{\theta=h_{\psi}(\eta),\bm{x}=h_{\phi}(\bm{\eps})}\\
	&= \int p(\eta) p(\bm{\eps}) \nabla_{\psi,\phi}q_{\phi}( \bm{a} | \theta,\bm{x})\\ 
	&\phantom{=}\cdot \log \hat{Z}_M^{\theta,\phi}  d(\eta, \bm{a},\bm{\eps}) \bigg|_{\theta=h_{\psi}(\eta),\bm{x}=h_{\phi}(\bm{\eps})}\\
	&={\E}_{p(\eta)p(\bm{\eps})q_{\phi}(\bm{a}| h_{\psi}(\eta),h_{\phi}(\bm{\eps}))} \Bigg[ \nabla_{\psi,\phi}\log\hat{Z}_M^{h_{\psi}(\eta),\phi} \\
	&\quad +\nabla_{\psi,\phi} \log q_{\phi}(\bm{a}| h_{\psi}(\eta),h_{\phi}(\bm{\eps})) \log \hat{Z}_M^{h_{\psi}(\eta),\phi} \Bigg].
\end{align*}
Analogously to \citet{maddison2017filtering,le2017auto,naesseth2017variational} in a variational EM framework, we have also ignored  the second summand in the gradient due to its high variance in our experiments. We take Monte Carlo samples of the expectation above and optimize the bound using Adam \citep{kingma2014adam}. It is also possible to use natural gradients \citep{amari1998natural}, see Appendix \ref{AppendixNaturalGradients}.

\section{Experiments}

\subsection{Linear Gaussian state space models}

\begin{figure}[htb]
	\begin{subfigure}{.23\textwidth}
		\centering
		\includegraphics[width=1.0\linewidth]{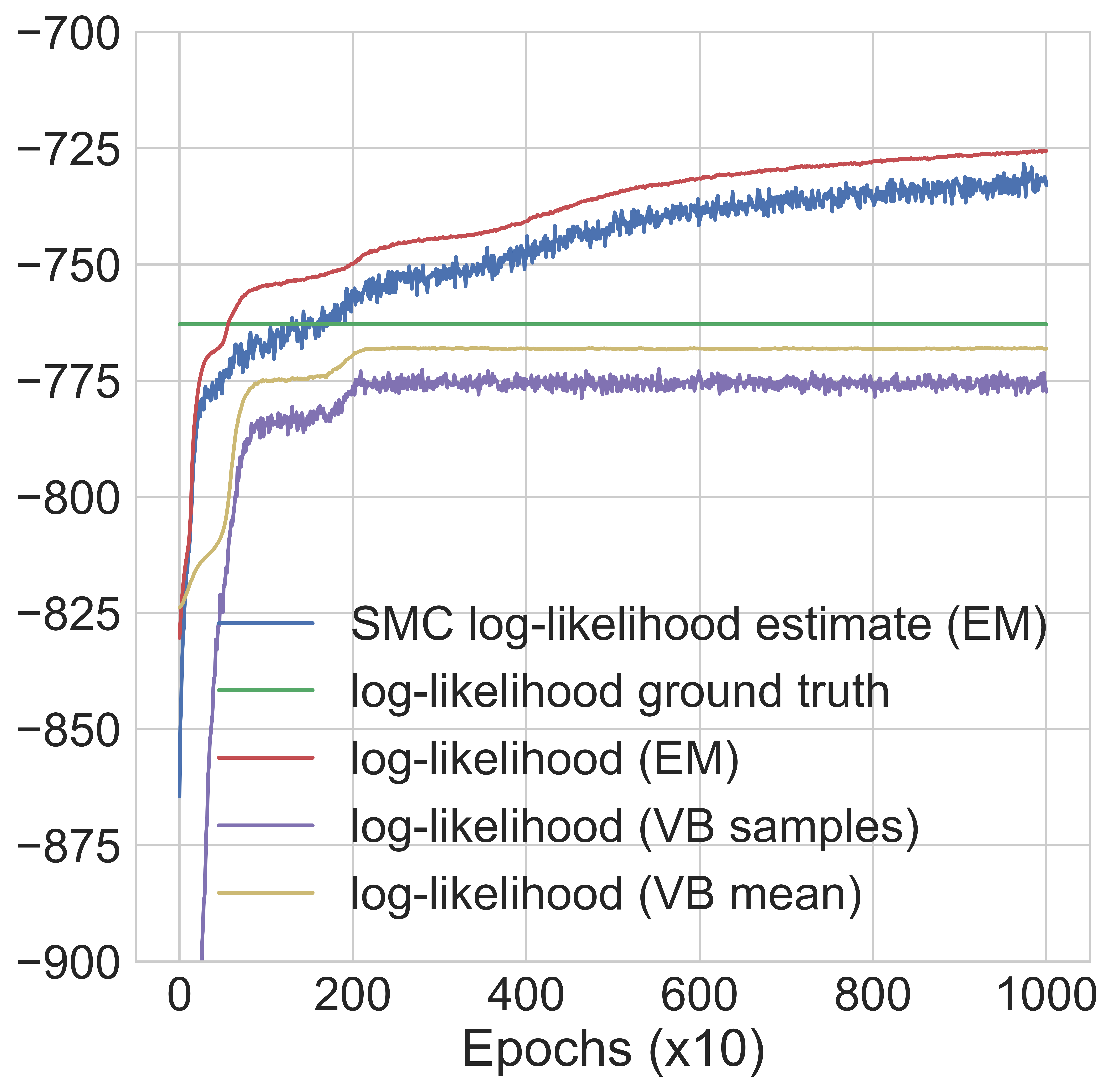}
		\caption{Log-likelihood on training data.}
		\label{LGSS_LLH_10_train}
	\end{subfigure}	
	\begin{subfigure}{.23\textwidth}
		\centering
		\includegraphics[width=1.0\linewidth]{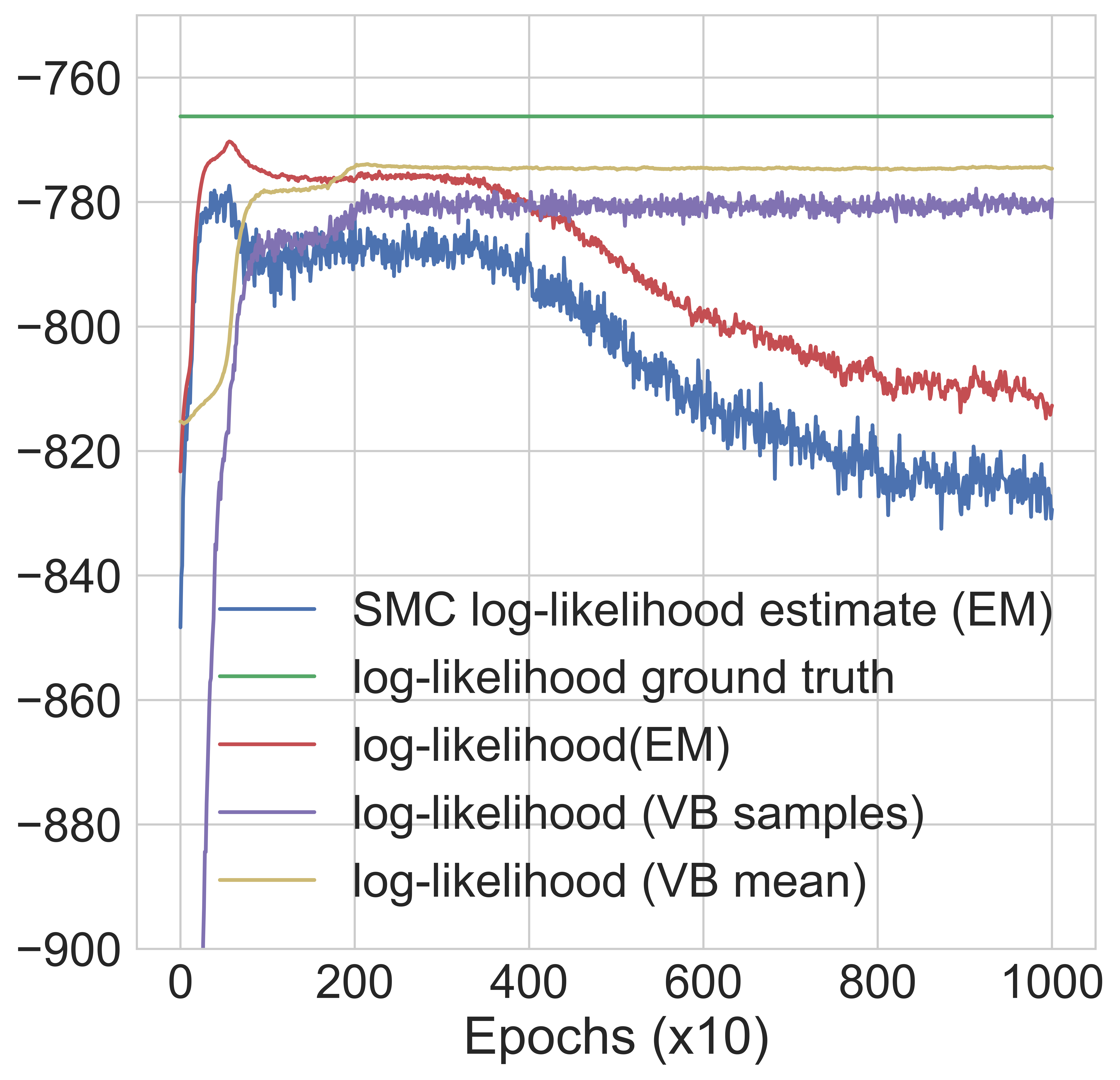}
		\caption{Log-likelihood on testing data.}
		\label{LGSS_LLH_10_test}
	\end{subfigure}	
	\caption{Log-likelihood for linear Gaussian state space models. Log-likelihood values are computed using Kalman filtering. The static parameters used in the VB case are the mean of the variational distribution (VB mean) or the samples from the variational distribution (VB samples) as they are drawn during training.}
	\label{LGSS_LLH}
\end{figure}

\paragraph{Regularisation in a high-dimensional model.}
We illustrate potential benefits of a fully Bayesian approach in a standard linear Gaussian state space model
\begin{align}
	f_{\theta}(x_{n}| x_{n-1})&=\mathcal{N}(A x_{n-1},\Sigma_x),\label{lgss_state_transition}\\
	g_{\theta}(y_n| x_n)&=\mathcal{N}(Bx_n,\Sigma_y)\label{lgss_observation},
\end{align}
with initial state distribution $X_0\sim \mathcal{N}(A^0,\Sigma_x^0)$
and parameters $A,\Sigma_x, \Sigma_x^0\in  \mathbb{R}^{d_x\times d_x}$, $A^0\in \mathbb{R}^{d_x}$, $B \in \mathbb{R}^{d_x \times d_y}$, and $C,\Sigma_y \in \mathbb{R}^{d_x \times d_y}$. 
\citet{naesseth2017variational} have shown in a linear Gaussian model that learning the proposal yields a higher variational lower bound compared to proposing from the prior and the variational bound is close to the true log-marginal likelihood for both sparse and dense emission matrices $B$. However, an EM approach might easily over-fit, unless one employs some regularisation, such as stopping early if the variational bound decreases on some test set. We demonstrate this effect by re-examining one of the experiments in \citet{naesseth2017variational}, setting $(d_x,d_y)=(10,3)$, $M=10$ and assume that $\Sigma_x$, $\Sigma_x^0$ and $\Sigma_y$ are all identity matrices. Furthermore, $A^0=0$ and $(A_{ij})=\alpha^{|i-j|+1}$ with $\alpha=0.42$, and $B$ has randomly generated elements with $B_{ij}\sim \mathcal{N}(0,1)$.
We assume that the proposal density is 
\[M_{n+1}^{\phi}(x_{n+1}| x_{n},y_{n+1})=\mathcal{N}(x_{n+1}|  A_{\phi}x_n+B_{\phi}y_{n+1},\Sigma_{\phi}),\]
and $M_{0}^{\phi}(x_{0}| y_{0})=\mathcal{N}(x_{0}|  A^0_{\phi}+B_{\phi}y_{0},\Sigma^0_{\phi})$, with $\Sigma_{\phi}$ and $\Sigma^0_{\phi}$ diagonal matrices. 
We perform both a variational EM approach and a fully Bayesian approach over the static parameters using $K=4$ particles. In the latter case, we place Normal priors $B_{ij}\sim \mathcal{N}(0,10)$ and $A_{ij}\sim \mathcal{N}(0,1)$. Furthermore, we suppose that a priori $ \Sigma_y$ is diagonal with variances drawn independently from an Inverse Gamma distribution with shape and scale parameters of $0.01$ each. A mean-field approximation for the static parameters is assumed. We suppose that the variational distribution over each element of  $A$ and $B$ is a normal distribution and the approximation over the diagonal elements of $\Sigma_y$ is log-normal. For identifiability reasons, we assume that $\Sigma_x$, $\Sigma_x^0$ and $A^0$ are known. We compare the EM and VB approach in terms of log-likelihoods on out-of-sample data assuming training and testing on $10$ iid sequences. Figure \ref{LGSS_LLH} shows that in contrast to the VB approach, the EM approach attains a higher log-likelihood on the training data with a lower log-likelihood on the test set as the training progresses.\\

\begin{figure}[htb]
	\begin{subfigure}{.23\textwidth}
		\centering
		\includegraphics[width=1.0\linewidth]{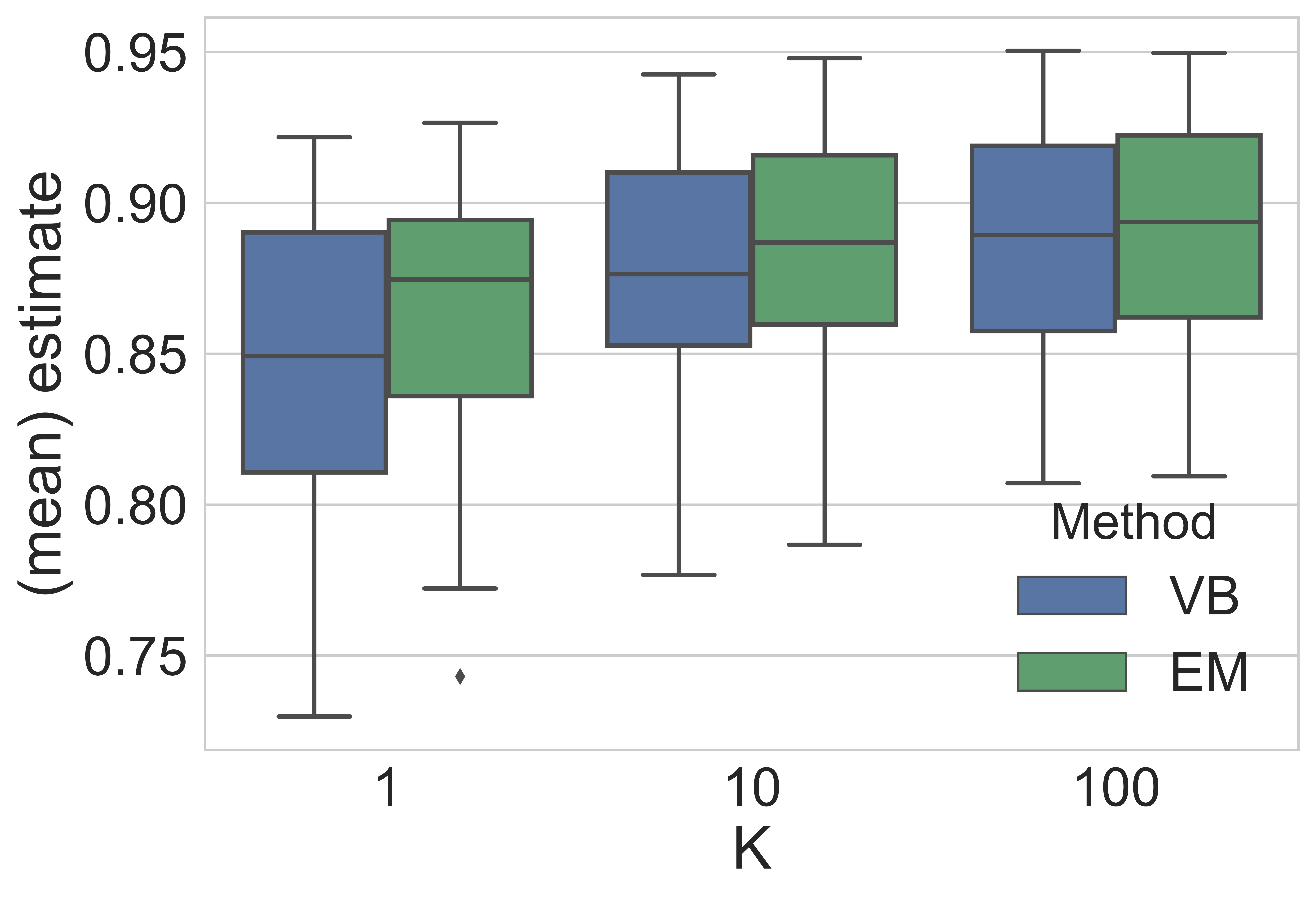}
		\caption{Point estimate of the autoregressive parameter $\lambda$ in the EM case or the variational mean in the VB case over $30$ simulations for $K\in\{1,10,100\}$ particles.}
		\label{LGSS_Boxplot}
	\end{subfigure}	
	\begin{subfigure}{.23\textwidth}
		\centering
		\includegraphics[width=1.0\linewidth]{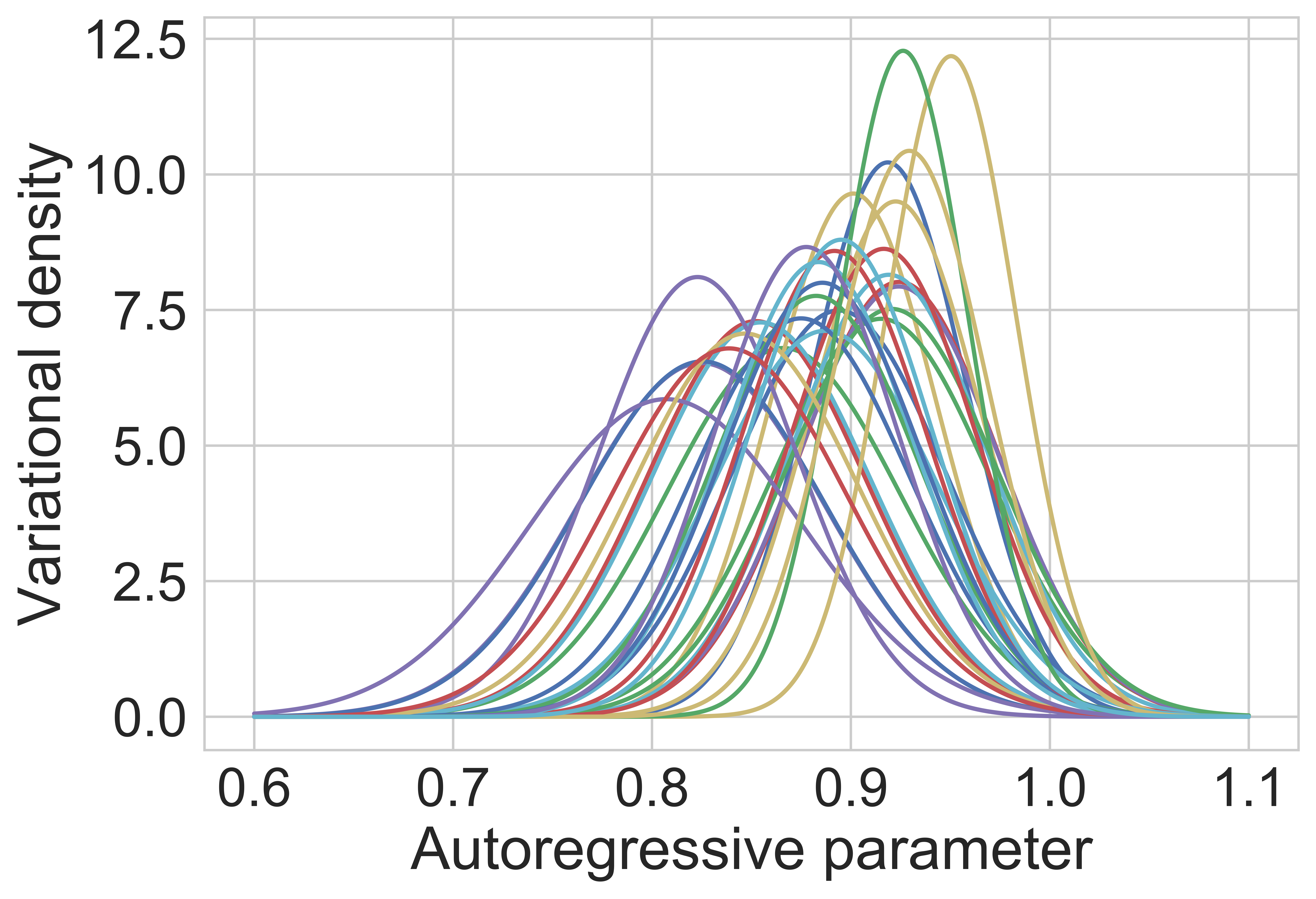}
		\caption{Variational distribution of the autoregressive parameter $\lambda$ using $K=100$ particles for each of the $30$ simulations.}
		\label{LGSS_AR_Variaional}
	\end{subfigure}	
	
	\caption{Inference on the autoregressive parameter $\lambda$ over $30$ simulations of length $M=100$. Ground truth values are $\lambda=0.9$.}
	\label{LGSS_AR_param}
\end{figure}

\paragraph{Approximation bias in a low-dimensional model.}
Variational approximations for the latent path can yield biased estimates of the static parameters, see \citet{turner2011two}. We illustrate that this bias decreases for increasing $K$ in a two-dimensional linear Gaussian model, both in an EM and VB setting.
We therefore consider inference in a linear Gaussian state space model (\ref{lgss_state_transition}-\ref{lgss_observation}) with two-dimensional latent states and one-dimensional observations. The state transition matrix is assumed to be determined by the autoregressive parameter $\lambda$ with $A=\begin{pmatrix}
\lambda & 0 \\ 0 & \lambda
\end{pmatrix}.$ We consider inference over $\lambda$ as the static parameter and fix $B=(1,1)$ with $\Sigma_x$ and $\Sigma_y$ being identity matrices. We simulate $30$ realisations of length $M=100$ each using $\lambda=0.9$. Inference is performed with different initialisations and learning rates over the simulated datasets. It has been documented in such a linear Gaussian model, see \citet{turner2011two}, that Gaussian variational approximations of the latent path that factorise over the state components underestimate $\lambda$. We observe the same effect in Figure \ref{LGSS_Boxplot} when using just $K=1$ particle. However, increasing the number of particles used during inference reduces this bias. 
Furthermore, we find that point estimates of the static parameters show some variation over different simulations, while an approximate Bayesian approach can be argued to better account for this uncertainty. 
The variational distributions for $\theta$ for each of the simulations using $K=100$ particles is shown in Figure \ref{LGSS_AR_Variaional}, confirming that they all put significant mass on the ground truth. Let us remark that these experiments also complement those in \citet{le2017auto}, where it is illustrated that increasing $K$ improves learning point estimates of the static parameters in a Gaussian model with a one-dimensional latent state. Indeed, as shown next, the marginal variational distribution allows not just for dependencies in the latent states across time, but also across different state dimensions, even if they are independent under the proposal.

\begin{figure}[htb]
	\begin{subfigure}{.23\textwidth}
		\centering
		\includegraphics[width=1.0\linewidth]{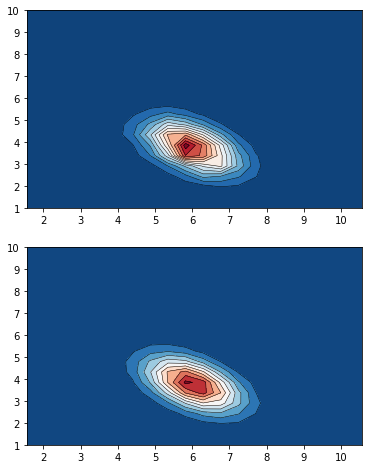}
		\caption{Joint distribution of the latent states at the second time step. Top: variational approximation, bottom: true posterior.}
		\label{csmct1_text}
	\end{subfigure}	
	\begin{subfigure}{.23\textwidth}
		\centering
		\includegraphics[width=1.0\linewidth]{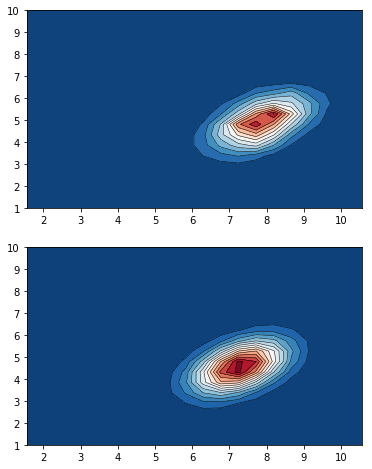}
		\caption{Joint distribution of the first state component at the first and second time step. Top: variational approximation, bottom: true posterior.}
		\label{csmcc1_text}
	\end{subfigure}

	\caption{Two-dimensional contour plots of the distribution of the latent path over two time steps and two state components. Function arguments are set to the ground truth state values as simulated if they are not shown.}
	\label{csmc_vis_text}
\end{figure}

\paragraph{Marginal variational distribution in a low-dimensional model.}
In an additional experiment, we evaluate if the variational approximation from Proposition $\ref{proposition_marginal_distribution}$ of the latent path matches the distribution of its true posterior. We consider the above state space model over $2$ time steps as in \citet{turner2011two}. Note that for given static parameters, the posterior is Gaussian. Indeed, for $\bm{x}=(x_{0}^{(0)},x_{0}^{(1)},x_{1}^{(0)},x_{1}^{(1)})$, where $x_{n}^{(i)}$ denotes dimension $i$ of $x_n$, we have $p(\bm{x}|y_{0:1},\lambda)=\mathcal{N}(\mu_{x|y},\Sigma_{x|y})$ with
\[\Sigma_{x|y}^{-1}=
\begin{pmatrix}
2&1&-\lambda& 0 \\
1&2&0&-\lambda\\
-\lambda& 0& 2&1\\
0&-\lambda &1 &2 &\\
\end{pmatrix},  
\mu_{x|y}=\Sigma_{x|y}\begin{pmatrix}
y_0\\y_0\\y_1\\y_1\end{pmatrix}, \]
assuming $X_0\sim \mathcal{N}(0,\frac{1}{1-\lambda^2} I)$ is drawn from its stationary distribution. We visualise the posterior distribution along with the marginal variational distribution 
\begin{align*}
&q_{\phi}(x^l_{0:M}| \theta)\\=&\gamma_{\theta}(x^l_{0:M}) {\E}_{\tilde{\pi}_{\text{CSMC}}(x_{0:M}^{\neg b_{0:M}^l},a_{0:M-1}^{\neg b_{0:M-1}^l} | \theta, x_{0:M}^l)}\left[\left({\hat{Z}^{\theta,\phi}_M}\right)^{-1}\right]
\end{align*}
in Figure \ref{csmc_vis_text} using $K=100$ particles and $50$ samples for the expectation. We find that the approximation mirrors the true posterior. In particular, it accounts for explaining-away between different dimensions of the latent state, although we have used isotropic proposals.

\subsection{Stochastic volatility models}
To show that our method allows inference of latent states and static parameters of higher dimensions, we consider a multivariate stochastic volatility model,
\begin{align*} f_{\theta}(x_n| x_{n-1})&=\mathcal{N}(\mu+\text{diag}(a)(x_{n-1}-\mu),\Sigma_x),\\
	g_{\theta}(y_n| x_n)&= \mathcal{N}(0,\exp(\text{diag}(x_n)),
\end{align*}
where $X_0\sim \mathcal{N}(\mu,\Sigma_x^0)$ with $x_n,y_n,\mu,a\in \mathbb{R}^D$, and covariance matrix $\Sigma_x\in \mathbb{R}^{D\times D}$, $\theta=(\mu,a,\Sigma_x,\Sigma_x^0)$. This model has been considered in \citet{guarniero2017iterated} using particle MCMC methods under the restriction that $\Sigma_x$ is band-diagonal to reduce the number of parameters. It is also more general than that entertained in \citet{naesseth2017variational} with $\Sigma_x$ assumed diagonal, see also \citet{chib2009multivariate} for a review on stochastic volatility models. We consider a fully Bayesian treatment as in \citet{guarniero2017iterated}, applied to the same data set of $90$ monthly returns (9/2008 to 2/2016) of $20$ exchange rates with respect to the US dollar as reported by the Federal Reserve System. The specification of the prior and variational forms of the static parameters are explained in Appendix \ref{appendix_stoch_vol_approximations}. We consider proposals of the form
\[M_{\phi}(x_{n+1}| y_{n+1},x_n)= \mathcal{N}(\mu+\text{diag}(a)(x_{n}-\mu),\Sigma^{\phi}),\] 
where $\Sigma^{\phi}$ is diagonal and using $K=50$ particles. Densities of the variational approximation that correspond to the GBP exchange rate can be found in Appendix \ref{appendix_stoch_vol_approximations}, Figure \ref{GBP_variational}, which are largely similar to those obtained in \citep{guarniero2017iterated}. Furthermore, we approximate the one- and two-step predictive distributions
\[p(y_{m+p}| y_{0:m}) \approx \frac{1}{S}\sum_{s=1}^S \sum_{k=1}^K W_m^{k,s} \delta_{X^{k,s}_{m+p}}p_{\theta_s}(y_{m+p}| X^{k,s}_{m+p})  \]
for $p\in \{1,2\}$,where $\theta_1,...,\theta_S \sim q_{\psi}(\theta)$, $\sum_{k=1}^K W_{m}^{k,s}\delta_{X_{m}^k}$ is the approximation of $p_{\theta_s}(x_m| y_{0:m})$ by the particle filter and $X_n^s\sim p_{\theta_s}(x_n^{k,s}| X_{n-1}^{s},Y^{k,s}_{n-1})$ with $Y^s_{n} \sim p_{\theta_s}(y^{k,s}_{n}| X^{k,s}_{n})$ for $n=m+1,...,m+p$ simulated from the generative model. The predictive distributions are evaluated using a log scoring rule \citep{gneiting2007strictly,geweke2010comparing} to arrive at the predictive log-likelihoods in Table \ref{pred_llh_table}. The full variational approach attains higher predictive log-likelihoods.

\begin{table}
	\caption{Average $p$-step predictive log-likelihoods per observation for the stochastic volatility model with different number of particles $K$ and number of samples $S$ from the variational distribution. In the EM case, we run $S$ particle filters with the same optimal static values. Mean estimates with standard deviation in parentheses based on 100 replicates.}
	\label{pred_llh_table}
	\centering
	\begin{tabular}{lll}
		\toprule
		&\multicolumn{2}{c}{$(S,K)=(4,50)$}\\		
		Method & $p=1$     & $p=2$  \\
		\midrule
		EM & 9.697 (0.008)  & 9.716 (0.008)       \\
		VB     & 9.707 (0.011) & 9.728 (0.015)       \\
		\midrule[0.25ex]
		&\multicolumn{2}{c}{$(S,K)=(20,100)$}\\		
		Method & $p=1$     & $p=2$  \\
		\midrule
		EM  & 9.690 (0.003)  & 9.713 (0.003)      \\
		VB  & 9.701 (0.004)  & 9.727 (0.005)    \\
		\bottomrule
	\end{tabular}
\end{table}

\subsection{Non-linear stochastic Hawkes processes}

There has been an increasing interest in modelling asynchronous sequential data using point processes in various domains, including social networks \citep{linderman2014discovering,wang2017linking}, finance \citep{bacry2015hawkes}, and electronic health \citep{lian2015multitask}. Recent work \citep{du2016recurrent,mei2017neural,xiao2017joint,xiao2017wasserstein} have advocated the use of neural networks in a black-box treatment of point process dynamics.

We illustrate that our approach allows scalable probabilistic inference for continuous-time event data $\{T_n,C_n\}_{n>0}$, $T_n<T_{n+1}$, where $T_n$ is the time when the $n$-th event occurs and $C_n\in \{1,...,D\}$ is an additional discrete mark associated with the event. We consider describing such a realisation as a $D$-variate point process with intensities $\lambda_{t}=h_{\theta}(\mu+\sum_{b=1}^B \Xi^b_t)$, driven by $B$ continuous time processes \[\Xi_t^{b}=\sum_{n\geq 1}  \beta_b A_n^b\e^{-\beta_b (t-T_n)}1_{\left[0,t\right)}(T_n),\quad t>0,\] 
and a non-negative monotone function $h_{\theta}$. Moreover, $\mu,A_n \in\mathbb{R}^D$ and $\beta^b>0$. Importantly, we allow $A_n^b$ to depend on $C_n$, and the $i$-th component of $A_n^b$ describes by how much the $n$-th event excites, if $(A_n^b)^i>0$, or inhibits, if $(A_n^b)^i<0$, subsequent events of type $i$. It is possible to view the dynamics as a discrete-time SSM; the essential idea being that $\Xi^b$ is piecewise-deterministic between events, see Appendix \ref{AppendixHawkesLiterature} for details along with related work on Hawkes point processes \citep{hawkes1971point}.
Let us define the discrete-time latent process $X_{n+1}=(Z_{n},A_{n})$ with $Z_n=\Xi_{T_n}$, $A_n=\text{vec}(A_n^1,...,A_n^B)$. Standard theory about point processes, see \citet{daleyintroduction1}, implies that the observation density is given by $g_{\theta}(t_n,c_n| z_{n-1})=\lambda_{t_n}^{c_n}\exp\left(-\sum_{i=1}^D\int_{t_{n-1}}^{t_n} \lambda_s^{i} ds\right)$,
where our model specification yields $\lambda_s$ as a deterministic function between $T_{n-1}$ and $T_n$ given $Z_{n-1}$. Similar to \citet{mei2017neural}, we set $h_{\theta}(y)=\nu \text{ softplus}(y/\nu)=\nu \log(1+\exp(y/\nu))$ as a scaled softplus function with $\nu$ a static parameter. Next, we specify the dynamics of $A_n$. We take the arguable most simple model, assuming $f_{\theta}(a_n| a_{n-1},z_{n-1},c_n)=\mathcal{N}(\sum_d \alpha_d \delta_{c_nd},\sum_d \sigma^2_d  \delta_{c_nd})$ with $\alpha_1,...,\alpha_D\in \mathbb{R}^{BD}$ and $\sigma^2_1,...,\sigma^2_D$ positive diagonal matrices, while remarking in passing that our approach allows readily for extensions that could include temporal dynamics between successive intensity jumps or intensity jumps instantaneously correlated across different marks and time scales. Due to the piecewise deterministic decay of $\Xi$, note that $Z_n^b| Z_{n-1}^b, A_n^b=\e^{-\beta^b(T_{n}-T_{n-1})}Z_{n-1}^b + \beta^b A_{n}^b$, so the state transition of the process $X$ is fully specified.\\ 
We apply our model to 20 days of high-frequency financial data for the BUND futures contract. The data is available as part of the tick library \citep{bacry2017tick} with $4$ event types: (i) mid-price up moves, (ii) mid-price down moves, (iii) buyer-initiated trades leaving the mid-price unchanged and (iv) seller-initiated trades not changing the mid. We train our model on 15 days and evaluate how well it predicts the type of the next event on out of sample data from the remaining 5 days.\\ 
Table \ref{hawkes_prediction_table} reports better predictive performance of the proposed model in comparison with two benchmark models. First, a linear Hawkes process model estimated using maximum likelihood. Second, to illustrate that improved predictions might not be just explained due to inhibitory effects, we also compare against a non-linear Hawkes model. The latter can be seen, and has been implemented, as a limiting case of our generative model letting $\sigma_d^2\to 0$, with inference thus performed using stochastic gradient descent of the negative log-likelihood.
Predictions are Monte Carlo samples of the next event realisation from the generative model. Further details including assumptions on the variational distributions and the predictive performance using a smaller training set are given in Appendix \ref{AppendixHawkesInference}.

\begin{table}
	\caption{Prediction metric for different Hawkes process models on the test set of around 206k events. The stochastic Hawkes model is trained with $20$ particles and uses $K\in\{20,80\}$ particles during testing.}
	\label{hawkes_prediction_table}
	\centering
	\begin{tabular}{ll}
	\toprule		
	Method & Error rate      \\
		&		next mark \\
	\midrule
	Linear Hawkes & 43.3 \%         \\
	Non-linear Hawkes & 40.9 \%         \\
	Non-linear stochastic Hawkes ($K=20$) &40.0\%        \\
	Non-linear stochastic Hawkes ($K=80$)& 39.3\%         \\	
	\bottomrule
	\end{tabular}
\end{table}

\section{Conclusion}

This paper has explored an inference approach that merges the scalability of variational methods with SMC sampling. 
We would like to emphasize that our approach is completely complementary to many recent advances in variational inference that can be used to parametrize $q_{\psi}(\theta)$. For instance, one can consider more expressive variational families \citep{rezende2015variational,kingma2016improved,salimans2015markov,maaloe2016auxiliary,ranganath2016hierarchical}. Similarly, our Bayesian approach naturally allows us to incorporate prior knowledge. For instance, one could place sparsity-inducing priors and impose corresponding variational approximations \citep{ingraham2017variational,ghosh2017model,louizos2017bayesian}. Applying such variational approximations to more expressive autoregressive models would be an interesting avenue to explore in future work.

\subsection*{Acknowledgements}
This research has been partly financed by the Alan Turing Institute under the EPSRC grant EP/N510129/1. The authors acknowledge the use of the UCL Legion High Performance Computing Facility (Legion@UCL), and associated support services, in the completion of this work.

\bibliography{bibliography_phd3}

\newpage

\appendix

\begin{appendices}
	
	\section{SMC algorithm}
	\label{AppendixSmcSampler}

	\begin{algorithm}
		\caption{Sampling from $q_{\phi}(x_{0:M}^{1:K},a_{0:M-1}^{1:K},l| \theta)$ via an SMC sampler}\label{SMCAlgo}
		\begin{algorithmic}[1]
			\State \textbf{Input}: observations $y_{0:M}$, prior density $p_{\theta}$, initial density $f_{\theta}(x_0)$, state transition density $f_{\theta}(x_{n+1}| x_n,y_n)$, observation density $g_{\theta}(y_n| x_n)$, proposal densities $M^{\phi}_n(x_n| y_n, x_{0:n-1})$ and resampling criteria.
			\State \textbf{Output}: $(X_{0:M}^{1:K},A^{1:K}_{0:M-1},L)\sim q_{\phi}(\cdot | \theta)$.
			\For{$k=1 ... K$}
			\State Sample $X_0^k\sim M_0^{\phi}(\cdot| y_0)$.
			\State Set $\alpha_0(X_0^k)=\frac{g_{\theta}(y_0| X_0^k) f_{\theta}(X_0^k| y_0)}{M_0^{\phi}(X_0^k)}.$
			\State Set $w_0(X_{0:n}^k)= \alpha_0(X^k_{0:n})/K$.
			\State Set $W_0^k\propto w_0(X_0^k).$
			\EndFor
			\For{$n=2 ... M$}
			\If{resampling criteria satisfied}
			\For{$k=1 ... K$}
			\State Sample $A_{n-1}^k\sim r(\cdot |W_{n-1}).$
			\EndFor
			\State Set $W_{n-1}=(\frac{1}{K},...,\frac{1}{K})$.
			\Else
			\State Set $A_{n-1}=(1,...,K)$.
			\EndIf
			\For{$k=1 ... K$}		
			\State Sample $X_n^k\sim M_n^{\phi}(\cdot | y_n, X_{0:n-1}^{A_{n-1}^k})$.
			\State Set $X_{0:n}^k=(X_{0:n-1}^k,X_n^k)$.
			\State Set $\alpha_n(X_{0:n}^k)=\frac{g_{\theta}(y_{n}| X_{n}^k)f_{\theta}(X_n^k| X_{n-1}^{A_{n-1}^k},y_{n-1})}{M_n^{\phi}(X_{n}^k| y_n, X_{0:n-1}^{A_{n-1}^k})}.$
			\State Set $w_n(X_{0:n}^k)=W_{n-1}^k \alpha_n(X^k_{0:n})$.
			\State Set $W_n^k \propto w_n(X_{0:n}^k)$.
			\EndFor
			
			\State Sample $L=l$ with probability $W^l_M$
			\EndFor
		\end{algorithmic}
	\end{algorithm}

	\section{Proof of Proposition \ref{proposition_extended_space}}
	\label{AppendixExtendedSpaceRepresentation}
	
	Consider an SMC algorithm with $K$ particles targeting 
	\[\pi_{\theta}(x_{0:M})\coloneqq \gamma(\theta,x_{0:M})/\gamma_M(\theta),\]
	where $\gamma(\theta,x_{0:M})=p(\theta,x_{0:M},y_{0:M})$ is related to the posterior via $\pi(\theta,x_{0:M})=\gamma(\theta,x_{0:M})/Z_M$. $Z_M$ is a normalising constant independent of $\theta$ that represents the marginal likelihood $Z_M=p(y_{0:M})$. Furthermore, $\gamma_M(\theta)=\int \gamma(\theta,x_{0:M})dx_{0:M}=p(\theta)p_{\theta}(y_{0:M})$. We denote the likelihood estimator of this SMC algorithm as $\tilde{Z}^{\theta,\phi}_M$.
	Following analogous arguments as in \citet{andrieu2010particle}, we have from the definition of the importance weights
	\begin{align*}
		&\frac{\tilde{\pi}(\theta,x_{0:M}^{1:K},a_{0:M-1}^{1:K},l)}{ q_{\phi,\psi}(\theta,x_{0:M}^{1:K},a_{0:M-1}^{1:K},l)}\\
		=&\frac{ 	
			\pi(\theta,x_{0:M}^l)K^{-(M+1)}}
		{q_{\psi}W_M^lM_0^{\phi}(x_0^{b_0^l}| y_0)\prod_{n=1}^M W_{n-1}^{b_{n-1}^l}M_{n}^{\phi}(x_n^{b_n^l} | y_n, x_{0:n-1}^{b_{n-1}^l})}\\		
		=&\frac{
			\pi(\theta,x_{0:M}^l)K^{-(M+1)} }
		{q_{\psi}(\theta)M_0^{\phi}(x_0^{b_0^l}| y_0)\prod_{n=1}^M M_{n}^{\phi}(x_n^{b_n^l} | y_n, x_{0:n-1}^{b_{n-1}^l}) } \\
		& \cdot \frac{\prod_{n=0}^M \left( \sum_{k=1}^K w_k(x_{0:M}^k)\right)}{\prod_{n=0}^M w_n(X_{0:M}^{b_n^l})}\\
		=&\frac{\pi(\theta,x_{0:M}^l)\tilde{Z}_M^{\theta,\phi}}{q_{\psi}(\theta)\gamma(\theta,x^l_{0:M})}\\
		=&\frac{\tilde{Z}_M^{\theta,\phi}}{q_{\psi}(\theta)p(y_{0:M})}. 
	\end{align*}	
	Note that $\tilde{Z}^{\theta,\phi}=p(\theta)\hat{Z}^{\theta,\phi}$, where $\hat{Z}^{\phi,\theta}$ is the SMC likelihood estimator in the main paper targeting a density proportional to $p_{\theta}(x_{0:M},y_{0:M})$, whilst $\tilde{Z}^{\theta,\phi}$ targets a density proportional to $p(\theta)p_{\theta}(x_{0:M},y_{0:M})$. Consequently,
	\begin{align*}\text{KL}(q_{\psi,\phi}||\tilde{\pi})&=-{\E}_{q_{\psi,\phi}}\left[\log \frac{\tilde{Z}_M^{\theta,\phi}}{q_{\psi}(\theta)}\right]+\log p(y_{0:M})\\
	=&-\mathcal{L}(\psi,\phi)+\log p(y_{0:M}),
	\end{align*}
	which concludes the proof.
	
	\section{Proof of Corollary \ref{corollary_marginal_bound}} \label{AppendixMarginalKL}
	Observe that we can write
	\begin{align*}&\text{KL}\left(q_{\psi,\phi}(\theta,x_{0:M}^{1:K},a_{0:M-1}^{1:K},l)||\tilde{\pi}(\theta,x_{0:M}^{1:K},a_{0:M-1}^{1:K},l)\right)\\
		&={\E}_{q_{\psi,\phi}(\theta,x_{0:M}^l,b_{0:M}^l)}\bigg[{\E}_{q_{\phi}(x_{0:M}^{\neg b_{0:M}^l},a_{0:M-1}^{\neg b_{0:M-1}^l})| \theta,x_{0:M}^l,b_{0:M}^l)}\Big[\\
		& \qquad  \qquad \log q_{\psi,\phi}(\theta,x_{0:M}^l,b_{0:M}^l) \\
		&\qquad \qquad  +\log q_{\phi}(x_{0:M}^{\neg b_{0:M}^l},a_{0:M-1}^{\neg b_{0:M-1}^l}| \theta,x_{0:M}^l,b_{0:M}^l) \bigg] \\
		&\quad \qquad -\log \tilde{\pi}(\theta,x_{0:M}^l,b_{0:M}^l)\\ & \quad \qquad -\log \tilde{\pi}_{\text{CSMC}}(x_{0:M}^{\neg b_{0:M}^l},a_{0:M-1}^{\neg b_{0:M-1}^l} | \theta,x_{0:M}^l,b_{0:M}^l)\ \bigg]\\
		&=\text{KL}(q_{\psi,\phi}(\theta,x^l_{0:M})||\pi(\theta,x^l_{0:M}))\\
		&\quad +{\E}_{q_{\psi,\phi}(\theta,x_{0:M}^l,b_{0:M}^l)}\bigg[\\
		& \qquad \qquad \text{KL}(q_{\phi}(x_{0:M}^{\neg b_{0:M}^l},a_{0:M-1}^{\neg b_{0:M-1}^l})| \theta,x_{0:M}^l,b_{0:M}^l)\Big|\Big|  \\
		& \qquad \qquad \qquad \tilde{\pi}_{\text{CSMC}}(x_{0:M}^{\neg b_{0:M}^l},a_{0:M-1}^{\neg b_{0:M-1}^l} | \theta,x_{0:M}^l,b_{0:M}^l))\bigg].
	\end{align*}
	
	\section{Proof of Proposition \ref{proposition_marginal_distribution}}	
	\label{AppendixMarginalDist}
	
	We can write the extended target distribution as
	\begin{align*} &\tilde{\pi}(x_{0:M}^{1:K},a_{0:M-1}^{1:K},l)\\=&\frac{\pi(\theta,x_{0:M}^l)}{K^{M+1}}\tilde{\pi}_{\text{CSMC}}(x_{0:M}^{\neg b_{0:M}^l},a_{0:M-1}^{\neg b_{0:M-1}^l} | \theta,x_{0:M}^l,b_{0:M}^l).
	\end{align*}
	This follows from the fact that $x_{0:M}^l=(x_0^{b_0^l},...,x_M^{b_M^l})$ and that $b_{0:M}| x_{0:M}^l,\theta$ is uniformly distributed on $\{1,...,K\}^{M+1}$. Hence, $\frac{\pi(\theta,x_{0:M}^l)}{K^{-(M+1)}}$ is the marginal density $\tilde{\pi}(\theta,x_{0:M}^l,b_{0:M}^l)$. Moreover, the variational approximation of the static parameter $\theta$ and latent states $x_{0:M}^l$, obtained as the marginal of the extended variational distribution,
	is given by, following similar arguments as in \citet{naesseth2017variational},
	\begin{align*}
		&q_{\psi,\phi}(\theta,x_{0:M}^l)=\frac{q_{\psi,\phi}(\theta,x_{0:M}^l,b_{0:M}^l)}{q_{\psi,\phi}(b_{0:M}^l| \theta,x_{0:M}^l)}\\
		&=\frac{1}{K^{-(M+1)}}\int q_{\psi,\phi}(\theta,x_{0:M}^l,a_{0:M-1}^l,x_{0:M}^{\neg b^l},a_{0:M-1}^{\neg b^l})\\
		&\phantom{\frac{1}{K^{-(M+1)}}\int}d(x_{0:M}^{\neg b^l},a_{0:M-1}^{\neg b^l})\\
		&=K^{M+1} \int q_{\psi}(\theta) \frac{w_M^l(x_{0:M}^{b^l})}{\sum_{l'}w_M^{l'}(x_{0:M}^{l'})} \prod_{k=1}^K M_0^{\phi}(x_0^k| y_0)\\
		&\phantom{K^{-(M+1)} \int } \cdot \prod_{n=1}^M \frac{w_{n-1}^k (x_{0:n}^{b_{n-1}^k})}{\sum_{l'}w_{n-1}^{l'}(x_{0:n-1}^{b_{n-1}^{l'}})} M_n^{\phi}(x_n^k| y_n,x_{0:n-1}^{b_{n-1}^{a_{n-1}^k}})\\
		&\phantom{K^{-(M+1)} \int } d(x_{0:M}^{\neg b^l},a_{0:M-1}^{\neg b^l})\\
		&=\int q_{\psi}(\theta) \left(\prod_{n=1}^M \frac{\gamma_{\theta}(x_{0:n}^l)}{\gamma_{\theta}(x^l_{0:n-1})\sum_{l'}w_n^{l'}((x_{0:n}^{l'}))}\right) \\
		&\phantom{\int}\cdot \prod_{k:k\neq b_0^l}M_0^{\phi}(x_0^k| y_0) \\
		&\phantom{\int}\cdot \prod_{n=1}^M \prod_{k:k\neq b_n^l} W_{n-1}^k M_n^{\phi}(x_n^k | y_n,x_{n-1}^{a_{n-1}^k}) d(x_{0:M}^{\neg b^l},a_{0:M-1}^{\neg b^l}) \\
		&=q_{\psi}(\theta)\gamma_{\theta}(x^l_{0:M})\\ &\quad \cdot {\E}_{\tilde{\pi}_{\text{CSMC}}(x_{0:M}^{\neg b_{0:M}^l},a_{0:M-1}^{\neg b_{0:M-1}^l} | \theta, x_{0:M}^l)}\left[\left({\hat{Z}^{\theta,\phi}_M}\right)^{-1}\right]
	\end{align*}

	\section{Natural gradients} \label{AppendixNaturalGradients}
	We have also experimented with optimizing the variational distribution over the static parameters using natural gradients \citep{amari1998natural,martens2014new} to take into account the Riemannian geometry of the approximating distributions, as explored previously for variational approximations, see for instance \citet{honkela2010approximate,hoffman2013stochastic}. Recall that we are optimizing over the space of probability distributions $q_{\psi}(\cdot)$ with parameter $\psi$, for which we can consider a possible metric given by the Fisher information
	\begin{align*}I(\psi)&={\E}_{q_{\psi}(\theta)}\left[\nabla_{\psi} \log q_{\psi}(\theta) \left(\nabla_{\psi} \log q_{\psi}(\theta)\right)^T\right]\\
	&=-{\E}_{q_{\psi}(\theta)}\left[H_{\log q_{\psi}}(\theta)\right], 
	\end{align*}
	The last equation assumes that $q_{\psi}$ is twice differentiable and ${H_{\log q_{\psi}}}(\theta)=\left(\frac{\partial^2 \log q_{\psi}(\theta)}{\partial \psi_i \partial \psi_j}\right)_{ij}$ denotes the Hessian. This induces an inner product $\langle \psi_1, \psi_2\rangle_{\psi_0}=\psi_1^T  F(\psi_0) \psi_2$ locally around $\psi_0$, hence gives rise to a norm $||\cdot ||_{\psi_0}$. The Fisher information matrix is connected to the KL divergence, since the distance in the induced metric is given approximately by the square root of twice the KL-divergence:
	\begin{align*} &\text{KL}(q_{\psi_1}||q_{\psi_2})\\=&\frac{1}{2}(\psi_2-\psi_1)I(\psi_1)(\psi_2-\psi_1)^T+O((\psi_2-\psi_1)^3),
	\end{align*}
	This follows from a second order Taylor expansion and from using the fact that ${\E}_{q_{\psi}}\left[\nabla_{\psi} \log q_{\psi}\right]=0$. Recall that the natural gradient of a function $\mathcal{L}(\psi)$ is defined by
	\[\tilde{\nabla}_{\psi}\mathcal{L}(\psi)=I(\psi)^{-1}\nabla_{\psi} \mathcal{L}(\psi)\]
	and one can show that under mild assumptions \citep{martens2014new},
	\begin{align*}&\sqrt{2}\frac{\tilde{\nabla}_{\psi}\mathcal{L}(\psi)}{||\tilde{\nabla}_{\psi}\mathcal{L}(\psi)||_{\psi}}\\=&\lim_{\eps\to 0}\frac{1}{\eps}\text{argmax}_{d:\text{KL}(q_{\psi+d}||q_{\psi})\leq \eps^2} \mathcal{L}(\psi+d).
	\end{align*}
	Thus the natural gradient is the steepest ascent direction with the distance measured by the KL-divergence. The natural gradient ascent does not depend on the parametrisation of $q_{\psi}$ as a consequence of the invariance of the KL-divergence with respect to reparametrisations.
	
	For mean-field approximations, computing the inverse of the Fisher information matrix simplifies, as the Fisher information has a block-diagonal structure in this case. We consider both normal and log-normal factors. For a univariate Gaussian distribution $q_{\mu,v}$ with mean $\mu$ and variance $\exp(v)^2$ parametrized by the logarithm of the standard deviation $v$, we obtain $\nabla_{\mu,v}\log q_{\mu,v}(\theta)=(\e^{-
		2v}(\theta-\mu),\e^{-2v}(\theta-\mu)^2-1)^T$. Consequently, 
	\[I(\mu,v)=\begin{pmatrix}
	\e^{-2v} & 0\\
	0 & 2
	\end{pmatrix}.\]
	For a log-normal distribution $q_{a,b}(\theta)$, parametrized so that $\log \theta\sim \mathcal{N}(a,\exp(b)^2)$, we have $\nabla_{a,b}\log q_{a,b}(\theta)=(\e^{-
		2b}(\log(\theta)-a),\e^{-2b}(\log(\theta)-a)^2-1)^T$ and we arrive at the same form for the Fisher information
	\[I(a,b)=\begin{pmatrix}
	\e^{-2b} & 0\\
	0 & 2
	\end{pmatrix}.\]

	\section{Priors and variational approximations for the stochastic volatility model}\label{appendix_stoch_vol_approximations}
	
\begin{figure}[htb]
		\begin{subfigure}{.23\textwidth}
			\centering
			\includegraphics[width=1.0\linewidth]{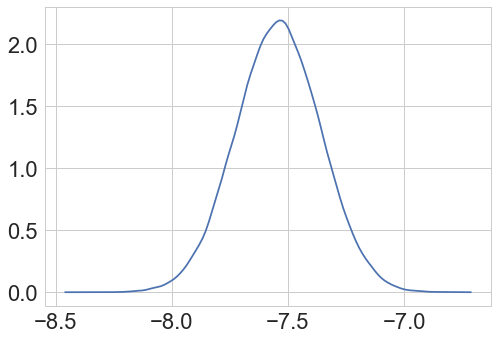}
			\caption{Mean reversion level $\mu$ of the log volatility related to the Pound Sterling.}
			\label{mu_GBP}
		\end{subfigure}	
		\begin{subfigure}{.23\textwidth}
			\centering
			\includegraphics[width=1.0\linewidth]{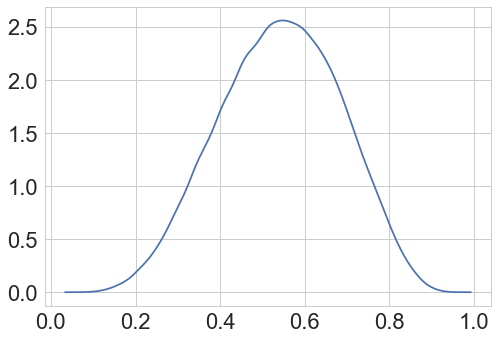}
			\caption{Autoregressive coefficient $a$ of the log volatility related to the Pound Sterling.}
			\label{AR_GBP}
		\end{subfigure}	
		\begin{subfigure}{.23\textwidth}
			\centering
			\includegraphics[width=1.0\linewidth]{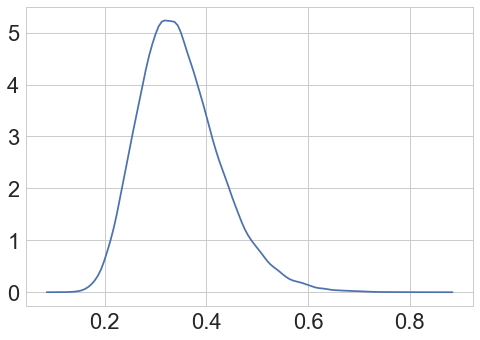}
			\caption{Variance part of $\Sigma_x$ for the error term of the log volatility related to the Pound Sterling.}
			\label{COV_GBP}
		\end{subfigure}	
		\begin{subfigure}{.23\textwidth}
			\centering
			\includegraphics[width=1.0\linewidth]{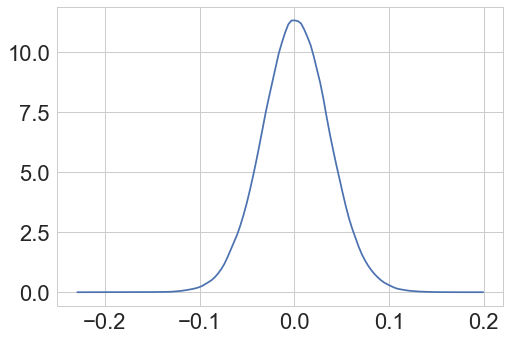}
			\caption{Covariance part of $\Sigma_x$ for the error term of the log volatilities related to the Pound Sterling and Euro.}
			\label{Cov_GBPEUR}
		\end{subfigure}	
		
		\caption{Density estimates for the parameters related to the Pound Sterling in the multivariate stochastic volatility model.}
		\label{GBP_variational}
	\end{figure}

	Compared to \citet{guarniero2017iterated}, we choose a different structure of $\Sigma_x$ to guarantee its positive-definiteness, along with slightly different priors. We model $\Sigma_x$ with its unique Cholesky factorisation \citep{dellaportas2012cholesky}, i.e. $\Sigma_x=LL^T$ with $L$ a lower triangular matrix having positive values on its diagonal. We set $\Sigma_x^0$ as the stationary covariance of the latent state. Independent priors are placed for $a_i\sim U(0,1)$ and $\mu_i\sim\mathcal{N}(0,10)$ as well as $L_{ij}\sim\mathcal{N}(0,10)$, for $i<j$ and $\log L_{ii}\sim \mathcal{N}(0,10)$. We assume a mean-field variational approximation with normal factors for $\mu$ and for the entries of $L$ below the diagonal and log-normal factors for its diagonal. Furthermore, $a_i$ is assumed to be the sigmoid transform sigm: $x\mapsto 1/(1+\e^{-x})$ of normally distributed variational factors. We initialized the mean of $L$ with a diagonal matrix having entries $0.2$ and the mean of $\mu_i$ with the logarithm of the standard deviation of the $i$th component of the time series. Densities of the variational approximation for parameters corresponding to the GBP exchange rate are given in Figure \ref{GBP_variational}.
	
	\section{Hawkes point processes and state space models}\label{AppendixHawkesLiterature}
	
	In contrast to linear Hawkes processes \citep{hawkes1971point,hawkes1971spectra}, we also allow for negative excitations, as explored previously for instance in \citet{bremaud1996stability,bowsher2007modelling,duarte2016stability}. 
	The values of $A^b$ and $\beta^b$ are commonly assumed to be fixed through time, while time-varying $\mu$ have been considered in various settings. 
	Stochastic time-varying excitations have been analysed in a probabilistic setting in \citet{bremaud2002power,dassios2011dynamic}. Moreover, \citet{ricci2014applied} considered frequentist inference of the excitation model parameters from a matrix-valued categorical distribution, while \citet{lee2016hawkes} performed MCMC with excitations evolving according to an Ito process in the one-dimensional case. However, scalable Bayesian inference for non-linear stochastic Hawkes processes has been missing, with previous variational inference schemes \citep{linderman2015scalable} having been restricted to linear Hawkes processes due to their resilience on the branching structure of linear Hawkes processes.
	SMC methods for shot-noise Cox processes has been considered in \citet{whiteley2011monte,martin2013inference} for on-line filtering and \citet{finke2014static} for static-parameter inference. While we expect such methods to scale poorly to models with many parameters and observations, we borrow their idea of describing the dynamics of the point process using piecewise-deterministic processes \citep{davis1984piecewise}, which enables us to employ the proposed inference approach for discrete-time state space models.
	\\
	More concretely, since $\Xi_t^b$ follows deterministic dynamics between two events, we can write $\Xi^b_t=F_b(t,T_n,\Xi_{T_n}^b)$ for $t\in \left[T_n,T_{n+1}\right)$ with the deterministic function 
	$F_{b}(t,s,z^b)={\e}^{-\beta_b(t-s)}z^b$. Whenever an event of type $C_n$ occurs at time $T_n$, the process $\Xi^b$ jumps with size $\Delta \Xi^b_{T_n}=\beta_b A_n^b$. The process $Z_n^b=\Xi_{T_n}^b$, $n >0$, satisfies $\Xi_t^b=F_b(t,T_n,Z_n^b)$ for $t\in \left[T_n,T_{n+1}\right)$. Note that we scale each $A_n^b$ with the diagonal matrix $\beta_b$. This ensures that the triggering kernel functions $s \mapsto  \beta^b \e^{-\beta^b s}$ have $L_0$ norm of one for any $b$.

	\section{Inference and predictions details for Hawkes process models}\label{AppendixHawkesInference}
	
	We place the following priors for the dynamics of $A$: For any $d\in \{1,...,D\}$, $\alpha_d\sim \otimes_{i=1}^{DB}\mathcal{N}(0,10)$ and consider mean-field variational approximations having the same forms. Furthermore, a priori, suppose that $\mu\sim \otimes_{i=1}^D \text{Ga}(0.01,0.01)$, $\text{diag}(\sigma_d^2)\sim \otimes_{i=1}^{DB}\text{Ga}(0.01,0.01)$ and $\beta_{b}-\beta_{b-1} \sim \mathcal{LN}(0,1)$, $b\in \{1,...,B\}, \beta_{0}=0$, all with a log-normal variational approximation. Eventually, for the softmax scale parameter, a priori $\nu \sim U(0,1)$ with a variational approximation as the sigmoid transform of a normal factor. The proposal function used is
	\begin{align} &M_{\phi}(a_n, z_n | a_{n-1},z_{n-1},t_{n+1},c_{n+1},t_n,c_n) \nonumber \\ 
	=&h_{\phi}(a_n| c_n)f_{\theta}(z_n| z_{n-1},a_{n-1},t_n,c_n),\label{hawkes_proposal}
	\end{align}
	with $h_{\phi}(a_n| c_n) =\mathcal{N}(\sum_d \tilde{\alpha}_d \delta_{c_nd},\sum_d \tilde{\sigma}^2_d \delta_{c_nd})$, $\tilde{\alpha_d} \in \mathbb{R}^{BD}$, $\tilde{\sigma_d}$ positive diagonal matrices and where$f_{\theta}$ describes the determinsitic decay of $Z_n$ according to the prior transition density. \\
	Let us also mention that the observation density contains a one-dimenisonal intractable integral. We apply Gaussian quadrature to evaluate the integral after transforming the quadrature points to
	better cover the interval immediately after an event where the intensity function is varying more
	quickly, see Appendix \ref{Quadrature} for details. We initialised the variational parameters so that the variational distribution of $\alpha$ is largely concentrated around the maximum likelihood estimates in a linear Hawkes model and the variational distribtuion of $\nu$ concentrated around $0$. The values of $\beta_b$ are commonly fixed in a maximum likelihood estimation setting to guarantee concavity of the log-likelihood. We have chosen $B=5$ with $(\log \beta_1,\log(\beta_2-\beta_1),...,\log(\beta_5-\beta_4))=(-1,1,3,5,7)$ fixed. This allows event interactions across various time scales, ranging from $\beta_1\approx 0.36$ to $\beta_5 \approx 1268$.\\ 
	We have also split the events in subsamples of length $M=100$ each and used 
	the particles from the previous event-batch as the initial particles for the subsequent event-batch. We used $K=20$ particles and performed optimisation with Adam \citep{kingma2014adam} and step size $0.0001$. Similar performance was observed either using standard or natural gradients for the considered hyperparameters and reported results correspond to optimsiaton with standard gradients only.\\
	
	Regarding inference for the benchmark models, maximum likelihood estimation for the linear Hawkes model was performed using the tick library \citep{bacry2017tick}, with the fixed time scales $\beta_1,...,\beta_5$ given above. Parameters for the non-linear Hawkes model were estimated using a limiting case of the generative model with very small $\sigma_d$, $K=1$, and proposing the single particle according to the generative model, hence particularly with small variances $\sigma_d$. Concretly, we consider
	\begin{align*}&f_{\theta}(a_n| a_{n-1},z_{n-1},c_n)\\=&h_{\phi}(a_n| c_n)=\mathcal{N}\left(\sum_d \alpha_d \delta_{c_nd}, \sum_d \sigma_d \delta_{c_nd} \right),
	\end{align*}
	recalling $h_{\phi}$ from the definition (\ref{hawkes_proposal}) of the proposal function and where for all $d\in \{1,...,D\}$,
	\[\sigma_d=\eps
	\begin{pmatrix}
	\beta_1^{-1}&&& & & \\
	& \ddots &\\
	& & \beta_1^{-1}\\
	& & &\ddots &\\
	& & & &\beta_B^{-1}\\
	& & & & &\ddots\\
	& & & & & & \beta_B^{-1}
	\end{pmatrix},\]
	$\eps=0.0001$. Stochastic gradient descent then yields point estimates over $\alpha_1,...,\alpha_D$, decay parameters $\beta_1,...,\beta_B$, softmax scale parameter $\nu$ and the background intensity parameter $\mu$. Initial parameters have similary been set to the maximum likelihood estimates from the linear Hawkes model. We used Adam \citep{kingma2014adam} with step sizes $0.0001$ and $0.0005$, with the reported result corresponding to the best performing step size for the considered metric in Table \ref{hawkes_prediction_table}. \\

	For the prediction of the next mark $c_{m+1}$ given the observations $t_{1:m},c_{1:m}$, we can sample $\theta_1,...,\theta_S\sim q_{\psi}(\theta)$ and run a particle filter that yields 
	\[\sum_{k=1}^K W_{m}^{k,s}\delta_{(Z_{0:m-1}^{k,s},A_{0:m-1}^{k,s})}(z_{0:m-1}^s,a_{0:m-1}^s)\]
	as an approximation of $p_{\theta_s}(z_{0:m-1}^s,\alpha_{0:m-1}^s| t_{1:m},c_{1:m})$. Set
	\[\hat{Z}_{m}^{b,k,s}=\e^{-\beta_b (t_{m}-t_{m-1})}Z_{m-1}^{b,k,s}+A_{m}^{b,k,s},\]
	with $A_{m}^{k,s} \sim f_{\theta_s}(\cdot| c_{m})$ sampled from the prior transition density.
	We then sample $10$ realisations
	\[t_{m+1}^{k,s,j},c_{m+1}^{k,s,j}\sim g_{\theta^s}(t_{m+1},c_{m+1}| \hat{Z}_m^{k,s}),\quad j=1,...,10,\] using the standard thinning algorithm for point processes, see for instance \citet{ogata1981lewis,daleyintroduction1,bowsher2007modelling}. In the stochastic Hawkes process model, we have chosen $S=4$ and $K=20$. To account for a similar computational budget for the benchmark models, we sample $10\cdot 4 \cdot 20$ event realisations in these cases instead.
	For predicting the next mark $c_{m+1}$, we use the sampled mark that occurred most often within $\{c_{m+1}^{k,s,j}\}_{k,s,j}$, where the count associated with $c_{m+1}^{k,s,j}$ is weighted by $W_{m}^{k,s}$.  Notice that we do not condition on the observed $t_{m+1}$ for predicting $c_{m+1}$ and the dependence of $c_{m+1}^{k,s,j}$ on $t_{m+1}^{k,s,j}$ is accounted for via the thinning procedure. In the stochastic Hawkes process model, we have also run predictions using $K=80$ particles, using the same model trained with $K=20$ particles. 
	
	In order to show how the different models generalize if less data is available, we have trained the different models on either the first 100 or 1000 events of one day and evaluated how well the model performs on predicting the first 10000 events on another day. We have repeated this procedure for 10 days and found that a fully Bayesian treatment is beneficial when trained on 100 events. The fully variational approach has an error rate of 65$\%$, whilst the same stochastic Hawkes process model using a point estimate of the static parameters has an error rate of 70$\%$. The two approaches yield similar results when trained on 1000 events with an error rate of below 50$\%$, whereas a benchmark non-linear Hawkes model without latent intensity dynamics has an error rate of 65$\%$. Although a fully Bayesian treatment might not be necessary if one imposes a parsimonious model for the evolution of the latent intensity, we hope that this example encourages further point process models that allow for online Bayesian updating as we feel that intensity excitations with latent dynamics have been underexplored for Hawkes process models.
	
	\section{Gaussian quadrature of the intensity function}\label{Quadrature}
	We approximate the integral of the intensity function with Gaussian quadrature, see for instance \citet{suli2003introduction} for details.
	Let $p_1,...,p_n$ be orthogonal polynomials in $L^2[a,b]$ equipped with the scalar product $\langle f,g\rangle=\int_a^b f(t)g(t)dt,$ $f,g\in L^2[a,b]$ with $p_k$ having degree $k$. Note that $p_k$ can be constructed recursively by Gram-Schmidt-orthogonalization. Furthermore, let $t_1,...,t_n$ be the roots of $p_n$ and consider the Lagrange polynomials for $i=1,...,n$,
	\[L_i(t)=\prod_{j=1,j\neq i}^n \frac{t-t_j}{t_i-t_j}, \]
	which satisfy $L_i(t_k)=\delta_{ik}, k=1,...,n$. Define 
	\[w_i=\int_a^bL_i(t)dt\]
	as well as the Gaussian quadrature 
	\[I_n(f)=\sum_{i=1}^nw_if(t_i).\]
	Then $I_n(p)=\int_a^b p(t)dt$ for polynomials $p$ of degree up to $2n-1$.
	We are interested in evaluating $\int_{T_{min}}^{T_{max}} \lambda^i(t)dt$ for fixed $T_{min}$ and $T_{max}$. Here, $T_{max}$ is the time of the next event and we have fixed $T_{min}$ to the previous event plus one microsecond. The lowest resolution of the event timestamps for the considered dataset is one microsecond. Assume there is a function $g$ such that $\lambda(t)=g(\e^t)$ . We can write 
	\[\int_{T_{min}}^{T_{max}}\lambda(t)dt=\int_{\log T_{min}}^{\log T_{max}} g(\e^{\tilde{t}})\e^{\tilde{t}}d\tilde{t}.\]
	This motivates the following change of variables that has also been considered in \citet{bacry2016estimation} for solving an integral equation involving the kernel function of a Hawkes process. Suppose that $t_1...t_n$ are the quadrature point with weights $w_1,...w_n$ on $[\log T_{min},\log T_{max}]$. The transformed quadrature scheme is then
	\[(\tilde{t}_n,\tilde{w}_n)=(\e^{t_n},w_n\e^{t_n}).\]
	We used $50$ quadrature points in our experiments.
	
\end{appendices}

\end{document}